\documentclass[sigconf,nonacm]{acmart}

\setcopyright{none}
\acmDOI{}
\acmISBN{}
\copyrightyear{2027}

\usepackage{graphicx}
\usepackage{amsmath}
\usepackage{booktabs}
\usepackage{multirow}
\usepackage{array}
\usepackage{xcolor}
\usepackage{colortbl}
\usepackage{placeins}
\usepackage{balance}

\definecolor{TableGroup}{gray}{0.97}
\newcolumntype{L}[1]{>{\raggedright\arraybackslash}p{#1}}
\newcolumntype{C}[1]{>{\centering\arraybackslash}p{#1}}
\providecommand{\uline}[1]{\underline{#1}}

\begin{document}

\title{DISRL: Dual-Scale Incremental Semi-Super\-vised Reinforcement Learning for Dynamic Community Tracking}

\author{Tianming Zhang}
\affiliation{
  \department{College of Computer Science and Software Engineering}
  \institution{Zhejiang University of Technology}
  \city{Hangzhou}
  \country{China}
}
\email{tmzhang@zjut.edu.cn}

\author{Jinxu Wang}
\affiliation{
  \department{College of Computer Science and Software Engineering}
  \institution{Zhejiang University of Technology}
  \city{Hangzhou}
  \country{China}
}
\email{jinxuwang@zjut.edu.cn}

\author{Yixuan Zhang}
\affiliation{
  \department{College of Computer Science and Software Engineering}
  \institution{Zhejiang University of Technology}
  \city{Hangzhou}
  \country{China}
}
\email{zhangyixuan@zjut.edu.cn}

\author{Zhengyi Yang}
\affiliation{
  \department{School of Computer Science}
  \institution{The University of Sydney}
  \city{Sydney}
  \country{Australia}
}
\email{zhengyi.yang@sydney.edu.au}

\author{Chenyu Hou}
\affiliation{
  \department{College of Computer Science and Software Engineering}
  \institution{Zhejiang University of Technology}
  \city{Hangzhou}
  \country{China}
}
\email{houcy@zjut.edu.cn}

\author{Bin Cao}
\affiliation{
  \department{College of Computer Science and Software Engineering}
  \institution{Zhejiang University of Technology}
  \city{Hangzhou}
  \country{China}
}
\email{bincao@zjut.edu.cn}

\author{Jing Fan}
\authornote{Corresponding author.}
\affiliation{
  \department{College of Computer Science and Software Engineering}
  \institution{Zhejiang University of Technology}
  \city{Hangzhou}
  \country{China}
}
\email{fanjing@zjut.edu.cn}

\renewcommand{\shortauthors}{Zhang et al.}


\begin{abstract}
Dynamic local community detection tracks the evolving community of designated seed nodes across graph snapshots. However, it faces two critical challenges: \textit{temporal supervision uncertainty}, where sparse and outdated labels fail to provide seed-specific guidance under topology changes, and \textit{unbounded error cascades}, where sequential policy transfer accumulates exploration errors over long horizons. To address these challenges, we propose DISRL, a Dual-scale Incremental Semi-supervised Reinforcement Learning framework. DISRL introduces a \textit{supervision construction module} that dynamically generates seed-specific guidance from sparse known communities, and a \textit{dual-scale community exploration module} that stabilizes policy evolution by decoupling short-term adaptation within temporal windows from long-term parameter consolidation across windows. Furthermore, we construct a dynamic graph benchmark with diverse community evolution scenarios. Extensive experiments demonstrate that DISRL consistently outperforms state-of-the-art baselines under various dynamic settings.

\end{abstract}

\ccsdesc[500]{Information systems~Social networks}
\ccsdesc[300]{Theory of computation~Graph algorithms analysis}
\keywords{Dynamic Graphs, Local Community Detection, Semi-supervised Learning, Reinforcement Learning}

\maketitle

\par

\section{Introduction}

Real-world networks, including social platforms~\cite{mazza-2023-modularity}, citation graphs~\cite{he-2023-h2cgl}, and fraud networks~\cite{wang-2024-dyhdge}, exhibit prominent community structures. In these networks, partially observed communities~\cite{zhang-2020-seal} often provide critical structural guidance for discovering seed-centered communities around entities of interest. Semi-supervised local community detection aims to leverage such sparse supervisory information to identify the specific community surrounding a seed node without exhaustively partitioning the entire graph~\cite{zafarmand-2023-fast}.
However, as interactions evolve across Web platforms and other real-world networks, communities surrounding queried users or entities may expand, shrink, merge, or split across temporal snapshots~\cite{paoletti-2025-codaen}, requiring adaptive approaches that can track local structural changes while preserving temporal coherence.


Existing methods remain inadequate for dynamic local community
tracking. Static approaches~\cite{zhang-2020-seal,wu-2022-clare,wu-2024-procom,ni-2024-semisupervised,xiong-2026-slfm,ni-2025-slrl,ni-2025-pretrained,kuang-2012-symmetric,he-2020-community,bhowmick-2024-dgcluster,blondel-2008-fast,holland-1983-stochastic}
ignore temporal dependencies by processing snapshots independently, while traditional dynamic
methods~\cite{mazza-2023-modularity,bakhtar-2025-devdynap,lin-2008-facetnet,yang-2011-detecting,chakrabarti-2006-evolutionary,zhuang-2021-dynamo} often rely on handcrafted
transition rules. Although recent learning-based models
~\cite{hashemi-2023-cstgn,costa-2023-ac2cd,kong-2024-learning,bianco-2025-semisupervised,liu-2024-deep,li-2023-dynamic,park-2022-cgc,zheng-2023-when,rossi-2020-temporal} capture complex temporal
patterns, they mainly process temporal streams passively via embedding aggregation rather than active exploration. Under
non-stationary evolution, such continual transfer may accumulate
tracking errors over long horizons.

To establish an active tracking paradigm, this paper investigates semi-supervised dynamic local community detection, aiming to accurately track the evolving community surrounding each seed node from graph snapshots with sparse supervision. This problem introduces two fundamental challenges.

\textit{First challenge: Temporal supervision uncertainty.} Rapid structural mutations (e.g., community splits or mergers) render historical predictions obsolete or misaligned with the current topology. Meanwhile, sparse supervision restricted to general known communities creates a structural mismatch with fine-grained, seed-specific local target boundaries.

\textit{Second challenge: Unbounded error cascade.} In non-stationary temporal graphs, the environmental Markov property is highly volatile. While continuously passing policy parameters across sequential snapshots preserves short-term temporal dependencies, exploratory errors from noisy snapshots can compound over long horizons, causing policy drift and performance degradation.

To address these challenges, we propose \textbf{DISRL}, a \textbf{D}ual-scale \textbf{I}ncremental \textbf{S}emi-Supervised \textbf{R}einforcement \textbf{L}earning framework that formulates dynamic local community tracking as structured sequential decision-making. DISRL combats these challenges via two synergistic modules. First, a \textit{semi-supervision construction module} resolves supervision uncertainty. By dynamically synthesizing seed-specific pseudo-reference communities from sparse annotations, it incrementally aligns historical predictions with current topologies, retrieving structurally relevant local guidance to counteract label sparsity. Second, a \textit{dual-scale community exploration module} mitigates unbounded error cascades through a decoupled, asynchronous policy adaptation pipeline. Within each temporal window (intra-window level), a fast policy propagates and adapts across consecutive snapshots to capture fine-grained evolutionary trajectories, initialized from a stable window-level master parameter. Concurrently, the master parameter remains stationary throughout the window to serve as a reliable geometric anchor and is centrally updated at the inter-window level only at the window's end by aggregating all intra-window transition gradients.

Furthermore, to bridge the evaluation gap caused by synthetic or oversimplified datasets, we construct a dynamic graph benchmark based on empirical network topologies and authentic community structures. It generates temporally ordered snapshots with diverse community evolution patterns, including expansion, shrinkage, splitting, merging, and topological perturbations, enabling rigorous evaluation under non-stationary settings. The main contributions are summarized as follows:
\begin{itemize}
    \item We pioneer the formulation of dynamic local community tracking as an incremental Reinforcement Learning (RL) sequence, shifting the traditional passive temporal embedding paradigms into a goal-directed, multi-step active exploration process around seed nodes.
    \item  We develop DISRL, featuring a semi-supervised supervision construction module that synthesizes seed-specific adaptive guidance, and an asynchronous dual-scale exploration pipeline that decouples intra-window trajectory optimization from inter-window gradient aggregation, effectively stabilizing policy transfer and mitigating the risk of long-horizon error accumulation.

    \item We construct a non-stationary dynamic graph
    benchmark that explicitly simulates fine-grained
    community evolutions.
    {Compared with the strongest baseline
    in each setting, DISRL achieves F1-score gains of
    2.89 and up to 6.42 percentage points on the
    real-world dataset and the simulated-evolution
    datasets, respectively.}
\end{itemize}

\section{Related Work}

\noindent\textbf{Static Community Detection:}
Static community detection methods have been extensively studied based on optimization~\cite{blondel-2008-fast}, matrix factorization~\cite{kuang-2012-symmetric}, generative models~\cite{holland-1983-stochastic}, and graph representation learning~\cite{perozzi-2014-deepwalk,grover-2016-node2vec,he-2020-community,bhowmick-2024-dgcluster}. To exploit limited prior knowledge, 
Semi-supervised methods are proposed. SEAL~\cite{zhang-2020-seal} adopts a generator-discriminator framework, SLSS~\cite{ni-2024-semisupervised} and SLFM~\cite{xiong-2026-slfm} exploit known communities to guide seed-centered local community detection. Recently, RL-based methods, including CLARE~\cite{wu-2022-clare} and SLRL~\cite{ni-2025-slrl}, formulate community expansion as sequential decision-making; prompt-based methods such as ProCom~\cite{wu-2024-procom} and PPSL~\cite{ni-2025-pretrained} improve label efficiency. However, these methods are designed for static topologies and cannot exploit temporal continuity, causing learned structural patterns to misalign with evolving communities.

\noindent\textbf{Dynamic Community Detection:}
Traditional approaches model temporal evolution through evolutionary clustering~\cite{chakrabarti-2006-evolutionary}, model-based approaches~\cite{lin-2008-facetnet,yang-2011-detecting}, incremental updates~\cite{zhuang-2021-dynamo}, seed-centered tracking such as DevDynaP~\cite{bakhtar-2025-devdynap}, and temporal community search methods~\cite{lin-2024-qtcs,zhou-2025-effective}. Recent learning-based methods leverage temporal graph neural networks~\cite{hashemi-2023-cstgn,li-2023-dynamic,zheng-2023-when,rossi-2020-temporal} and topology-regularized clustering~\cite{kong-2024-learning,liu-2024-deep,park-2022-cgc} to capture complex evolutionary patterns. 
Nevertheless, these methods rarely address seed-centered local community detection under sparse supervision. Without explicit seed-specific supervision, sparse known communities may provide weak guidance for evolving local targets; meanwhile, continual parameter transfer across snapshots can propagate transient errors over long horizons. These limitations call for an adaptive and stable framework for dynamic local community detection.

\section{Methodology}
\label{sec:methodology}

\subsection{Problem Formulation and Overview}
\label{sec:problem_overview}

Let $\mathcal{G}=\{G^t=(\mathcal{V}^t,\mathcal{E}^t)\}_{t=1}^{T}$ denote a dynamic graph with $T$ temporal snapshots, where $\mathcal{V}^t$ and $\mathcal{E}^t$ are the node and edge sets at timestamp $t$, respectively. At each snapshot, only a sparse set of known communities is available as incomplete supervision, denoted by $\mathcal{C}_{\mathrm{known}}^t=\{C_i^t\}_{i=1}^{m}$, where $C_i^t\subseteq\mathcal{V}^t$ ($1 \le i \le m$). Given a set of designated seed nodes $\mathcal{S}$, our goal is to track the evolving local community of each seed node over temporal snapshots. Specifically, for each seed $s\in\mathcal{S}$ at snapshot $G^t$, we aim to identify a community $C_{\mathrm{pred}}^t(s)\subseteq\mathcal{V}^t$ satisfying $s\in C_{\mathrm{pred}}^t(s)$. The objective is to sequentially infer the dynamic local communities $\mathcal{C}_{\mathrm{pred}}=\{C_{\mathrm{pred}}^t(s)\mid s\in\mathcal{S},t=1,\ldots,T\}$ by leveraging temporal graph evolution and sparse supervision.

\begin{figure*}[!t]
\centering
\includegraphics[width=1\textwidth]{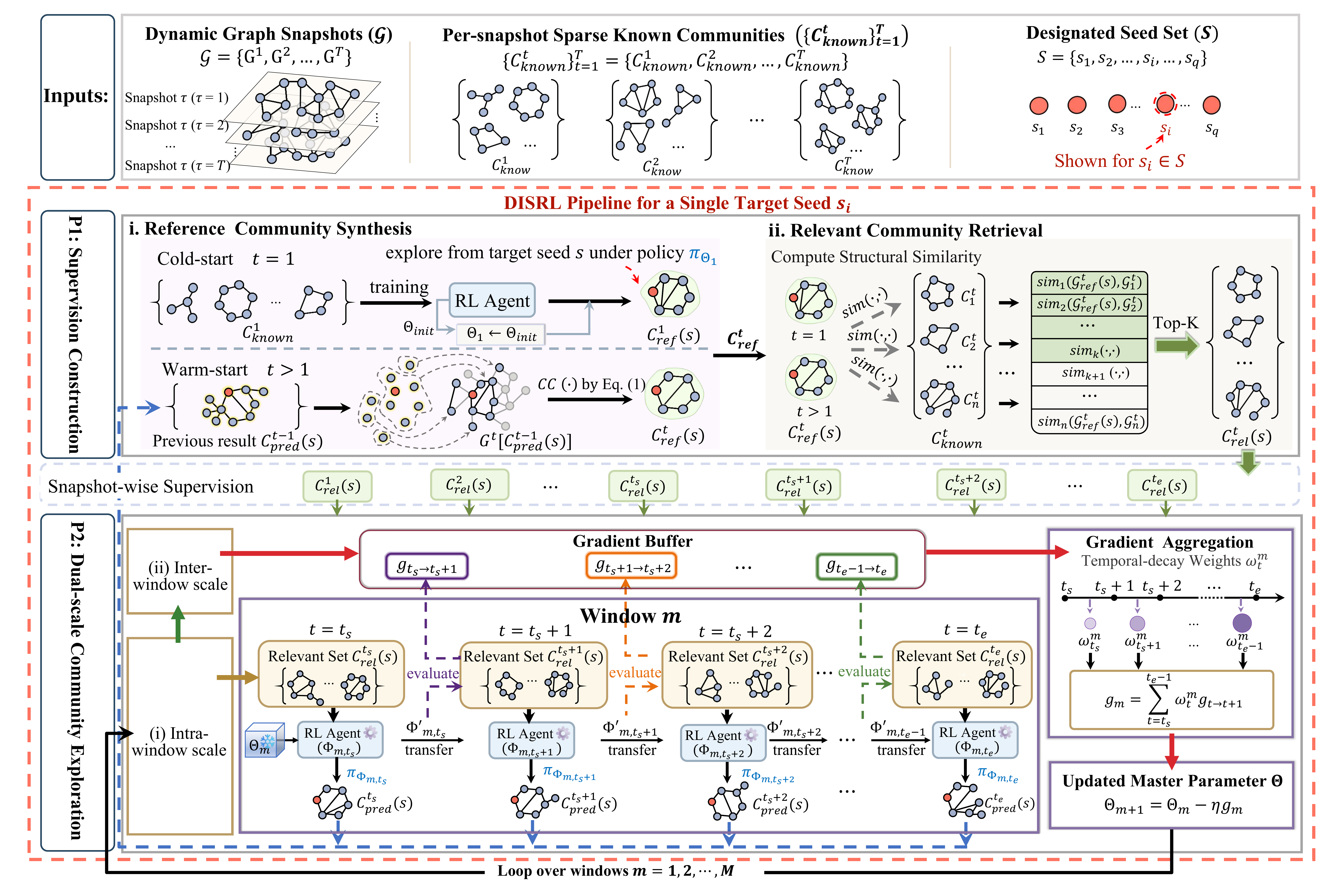}
\caption{Overall framework of DISRL, consisting of supervision construction and dual-scale community exploration.}
\Description{The framework passes seed-specific guidance to a community explorer. Fast policy parameters adapt over snapshots within each temporal window; a master parameter consolidates information between windows.}
\label{fig:framework}
\end{figure*}

Existing reinforcement learning-based community detection methods mainly target static graphs or short-term refinement, and cannot address \textit{temporal supervision uncertainty} and \textit{unbounded error cascade} in dynamic local tracking. To address these challenges, we propose DISRL, which is an incremental reinforcement learning framework.  As shown in Fig.~\ref{fig:framework}, DISRL consists of two synergistic modules. The \textit{Supervision Construction Module} resolves temporal supervision uncertainty by dynamically refining sparse known communities into seed-specific adaptive guidance $\mathcal{C}_{\mathrm{rel}}^t$. The \textit{Dual-scale Community Exploration Module} performs reinforcement-learning-based boundary expansion under this guidance. To mitigate long-horizon error cascades, DISRL decouples policy optimization into intra-window fast adaptation across consecutive snapshots and inter-window master anchoring across temporal windows.

\subsection{Supervision Construction Module}
\label{sec:supervision_construction}

In dynamic graphs, coarse global annotations $\mathcal{C}_{\mathrm{known}}^t$ capture macro-level partitions, often exhibiting spatial and structural mismatch with the micro-vicinity of a target seed $s$. To bridge this supervision gap without losing temporal consistency, the Supervision Construction module implements a two-stage process: reference community synthesis and relevant community retrieval.

\subsubsection{Reference Community Synthesis.}
The objective is to construct a topology-congruent local structural anchor $C_{\mathrm{ref}}^t$ for the current snapshot $G^t$. Rather than serving as the final prediction, this reference acts as an inductive bias to filter out irrelevant global annotations. For a specified seed $s$, $C_{\mathrm{ref}}^t(s)$ is defined as:
\begin{equation}
C_{\mathrm{ref}}^t(s) =\mathrm{CC}\left(G^t\big[C_{\mathrm{pred}}^{t-1}(s)\cap \mathcal{V}^t\big]\right)
\label{eq:reference_synthesis}
\end{equation}
where $G^t[V^\prime]$ is the subgraph of $G^t$ induced by $V^\prime$, and $\mathrm{CC}(\cdot)$ extracts the connected component containing $s$.

\noindent\textbf{\textit{Cold-start Bootstrapping ($t=1$):}} At the temporal onset, historical predictions are inaccessible ($t=1$). DISRL switches to the cold-start mode, executing the Community Explorer in the inference phase (as defined in Sec.~\ref{sec:scale_1_intra}) over $G^1$ initialized by a generic policy pre-trained on $\mathcal{C}_{\mathrm{known}}^1$. The generated initial seed-centric boundary is captured as $C_{\mathrm{ref}}^1$.

\noindent\textbf{\textit{Warm-start Alignment ($t>1$):}} For subsequent snapshots, directly inheriting $C_{\mathrm{pred}}^{t-1}(s)$ introduces geometric misalignment due to localized edge mutations in $G^t$. DISRL projects $C_{\mathrm{pred}}^{t-1}(s)$ onto the current topology. As formalized in Eq.~\eqref{eq:reference_synthesis}, valid active nodes are retained, while $\mathrm{CC}(\cdot)$ prunes disconnected fragments caused by community shrinkage or splits, ensuring that $C_{\mathrm{ref}}^t$ remains seed-connected.
\subsubsection{Relevant Community Retrieval.}
To eliminate macro-level topological noise, DISRL retrieves a compact set of known communities that are structurally similar to the synthesized reference community $C_{\mathrm{ref}}^t(s)$. Let $\mathcal{G}_{\mathrm{ref}}^t(s)=G^t[C_{\mathrm{ref}}^t(s)]$ denote the reference subgraph. For each known community $C_i^t\in\mathcal{C}_{\mathrm{known}}^t$, let $\mathcal{G}_i^t=G^t[C_i^t]$ denote its induced subgraph. To capture fine-grained path-level structural patterns without resorting to NP-hard exact graph isomorphism, DISRL employs a normalized shortest-path graph kernel~\cite{borgwardt-2005-shortestpath}. The structural similarity between the reference and the $i$-th known community is computed as:
\begin{equation}
\resizebox{0.90\columnwidth}{!}{$\displaystyle
\operatorname{sim}(\mathcal{G}_{\mathrm{ref}}^t(s),\mathcal{G}_i^t)=\frac{K_{\mathrm{sp}}(\mathcal{G}_{\mathrm{ref}}^t(s),\mathcal{G}_i^t)}{\sqrt{K_{\mathrm{sp}}(\mathcal{G}_{\mathrm{ref}}^t(s),\mathcal{G}_{\mathrm{ref}}^t(s))\cdot K_{\mathrm{sp}}(\mathcal{G}_i^t,\mathcal{G}_i^t)}},
$}
\label{eq:kernel_similarity}
\end{equation}
where $K_{\mathrm{sp}}(\cdot,\cdot)$ counts matching shortest-path lengths between graph node pairs, preserving geodesic path distributions that are sensitive to community boundary variations.

Subsequently, DISRL ranks all candidates in $\mathcal{C}_{\mathrm{known}}^t$ in descending order of the similarity scores in Eq.~\eqref{eq:kernel_similarity} and selects the top-$K$ communities to form the seed-specific relevant supervision set $\mathcal{C}_{\mathrm{rel}}^t$:
\begin{equation}
\resizebox{0.90\columnwidth}{!}{$\displaystyle
\mathcal{C}_{\mathrm{rel}}^t(s)=\operatorname{Top-K}\Bigl(\bigl\{C_i^t\in\mathcal{C}_{\mathrm{known}}^t\bigm|\operatorname{sim}(\mathcal{G}_{\mathrm{ref}}^t(s),\mathcal{G}_i^t)\bigr\}\Bigr).
$}
\label{eq:topk_selection}
\end{equation}

\subsection{Dual-scale Community Exploration Module}

\label{sec:dual_scale_exploration}

After obtaining $\mathcal{C}_{\mathrm{rel}}^t(s)$, DISRL formulates seed-centered community tracking on $G^t$ as an MDP-based sequential expansion process. As shown in Fig.~\ref{fig:framework}, the exploration module uses a dual-scale update pipeline to coordinate snapshot-level micro-exploration with temporal horizon-level macro-anchoring. At the intra-window scale, a fast policy $\Phi_{m,t}$ executes incremental transfer and rapid adaptation
across adjacent snapshots to capture short-term topology changes. At the inter-window scale, a master parameter $\Theta_m$ serves as a long-term stable anchor, periodically consolidating transition meta-gradients at window boundaries to limit the consecutive propagation of historical exploration errors and mitigate the risk of unbounded error cascades across long horizons.


\subsubsection{Intra-window Scale Exploration and Fast Adaptation.}
\label{sec:scale_1_intra}
To efficiently approximate target boundaries without global graph traversal, DISRL employs a Community Explorer based on an active search policy. During training, the policy is fine-tuned within $\mathcal{C}_{\mathrm{rel}}^t$: for each $C^*\in\mathcal{C}_{\mathrm{rel}}^t$, the agent performs rollouts from a randomly sampled virtual seed, with rewards derived from $C^*$. During inference, the adapted policy is deployed on the target seed $s$ to greedily explore the snapshot graph $G^t$ and obtain the predicted community $C_{\mathrm{pred}}^t(s)$. The exploration process is formulated as a snapshot-level MDP.

\noindent\textbf{\textit{State Space.}}
At exploration step $k$ in snapshot $G^t$, the environment state $\mathbf{S}_k^t$ is composed of (i) the currently explored community $C_k^t$; (ii) its immediate candidate action boundary $\mathcal{N}(C_k^t)$, which is defined as: 
\begin{equation}
\mathcal{N}(C_k^t)
=
\left\{
u \in \mathcal{V}^t \setminus C_k^t
\;\middle|\;
\exists\, v \in C_k^t,\ (u,v) \in \mathcal{E}^t
\right\},
\label{eq:boundary_definition}
\end{equation}
and (iii) the node representations computed on a subgraph centered around $C_k^t$. 

To capture the local structural context efficiently, DISRL extracts an $L$-hop localized subgraph induced by the expanded community region:
\begin{equation}
\mathcal{V}_{\mathrm{sub},k}^t
=
C_k^t \cup \mathcal{N}^L(C_k^t),
\quad
\mathcal{G}_{\mathrm{sub},k}^t
=
G^t\!\left[\mathcal{V}_{\mathrm{sub},k}^t\right],
\label{eq:local_subgraph}
\end{equation}
where $\mathcal{N}^L(C_k^t)$ is the $L$-hop neighborhood set enclosing $C_k^t$.

Within the subgraph $\mathcal{G}_{\mathrm{sub},k}^t$, we initialize a two-channel indicator matrix $\mathbf{X}_{k}^{t}\in\mathbb{R}^{|\mathcal{V}_{\mathrm{sub},k}^t|\times 2}$ to explicitly parameterize the geometric orientation of the root seed $s_{\mathrm{init}}$ and the membership of the currently explored community $C_k^t$. For each node $u \in \mathcal{V}_{\mathrm{sub},k}^t$, its corresponding row is defined as:
\begin{equation}
\mathbf{X}_{k}^{t}(u,:)
=
\left[
\mathbb{I}(u=s_{\mathrm{init}}),
\mathbb{I}(u\in C_k^t)
\right],
\label{eq:indicator_matrix}
\end{equation}
where $\mathbb{I}(\cdot)$ denotes the indicator function.

To effectively propagate the localized structural context and resolve the structural anonymity of partitions, an incremental Graph Pointer Network (iGPN)~\cite{zhang-2020-seal} is then deployed over $\mathcal{G}_{\mathrm{sub},k}^t$. Setting the base hidden state $\mathbf{H}^{(0)} = \mathbf{X}_{k}^{t}$, the latent node embeddings $\mathbf{H}^{(l)} \in \mathbb{R}^{|\mathcal{V}_{\mathrm{sub},k}^t|\times 2}$ at the $l$-th propagation layer are iteratively computed as:
\begin{equation}
\mathbf{H}^{(l)}
=
\alpha \widetilde{\mathbf{A}}_{\mathrm{sub},k}^t \mathbf{H}^{(l-1)}
+
(1-\alpha)\mathbf{X}_{k}^{t},
\quad l = 1,2,\dots,L
\label{eq:igpn_local}
\end{equation}
where $\widetilde{\mathbf{A}}_{\mathrm{sub},k}^t \in \mathbb{R}^{|\mathcal{V}_{\mathrm{sub},k}^t|\times |\mathcal{V}_{\mathrm{sub},k}^t|}$ is the symmetrically normalized adjacency matrix of $\mathcal{G}_{\mathrm{sub},k}^t$, and $\alpha \in (0,1)$ is the propagation coefficient. The final output $\mathbf{H}^{(L)}$ yields continuous node representations that simultaneously encapsulate seed-aware and community-aware features.

To strictly penalize boundary leakage and shape distortions during sequential expansion, we further introduce a community-level structural deviation feature $\delta_{u,k}^t \in \mathbb{R}$ for each node $u \in \mathcal{V}_{\mathrm{sub},k}^t$. 

\begin{equation}
\delta_{u,k}^t =
\Big| \mathbf{H}^{(L)}(u,2)
- \frac{1}{|C_k^t|}
\sum_{v\in C_k^t} \mathbf{H}^{(L)}(v,2)
\Big|.
\label{eq:deviation}
\end{equation}
Subsequently, the propagated structural feature channels and localized deviation scalar are concatenated into a compact low-dimensional node feature vector $\mathbf{z}_{u,k}^t \in \mathbb{R}^3$:
\begin{equation}
\mathbf{z}_{u,k}^t
=
\left[
\mathbf{H}^{(L)}(u,:)
\parallel
\delta_{u,k}^t
\right].
\label{eq:joint_vector}
\end{equation}
Finally, the complete environment state $\mathbf{S}_k^t$ is formalized as:
\begin{equation}
\mathbf{S}_k^t
=
\left(
C_k^t,
\mathcal{N}(C_k^t),
\left\{
\mathbf{z}_{v,k}^t
\;\middle|\;
v \in C_k^t \cup \mathcal{N}(C_k^t)
\right\}
\right).
\label{eq:encoded_state}
\end{equation}

\noindent\textbf{\textit{Action Space.}}
At exploration step $k$, the action space $\mathcal{A}_k^t$ consists of boundary candidates and a termination token ($\mathrm{EOS}$):
\begin{equation}
\mathcal{A}_k^t = \mathcal{N}(C_k^t) \cup \{\mathrm{EOS}\}.
\label{eq:action_space}
\end{equation}

Selecting $a = u \in \mathcal{N}(C_k^t)$ expands the current community as $C_{k+1}^t=C_k^t\cup\{u\}$, while $\mathrm{EOS}$ terminates exploration and returns the final community. 
To bound exploration, we additionally
terminate the rollout when the explored community
reaches the maximum size among the available known
communities.

The policy $\pi_{\Phi_{m,t}}(\cdot|\mathbf{S}_k^t)$ is parameterized by an MLP over node representations. During training, the fast policy is updated via policy gradient:
\begin{equation}
\Phi_{m,t}^{'}
=
\Phi_{m,t}
+
\eta
\sum_{k=0}^{T_e}
\nabla_{\Phi_{m,t}}
\log
\pi_{\Phi_{m,t}}
\left(a_k^t\mid\mathbf{S}_k^t\right)
R_k^t,
\label{eq:policy_gradient_update}
\end{equation}
where $\eta$ is the learning rate, $T_e$ is the terminal exploration
step, and
$R_k^t=\sum_{j=k}^{T_e}\gamma^{j-k}r_j^t$ is the discounted return
with discount factor $\gamma$ and immediate reward
$r_j^t=\operatorname{F1}(C_{j+1}^t,C^*)-
\operatorname{F1}(C_j^t,C^*)$.
The updated policy is then transferred to initialize the explorer at the next snapshot:
\begin{equation}
\Phi_{m,t+1} \leftarrow \Phi_{m,t}^{'}, \quad \forall\, t \in \{t_s, \dots, t_e-1\},
\label{eq:fast_parameter_transfer}
\end{equation}
  $t_s$ and $t_e$ are the start and end timestamps of the window, respectively. This incremental policy inheritance enables fast adaptation to local temporal changes while preserving evolutionary continuity within each temporal window.

\subsubsection{Inter-window Scale Meta-Anchoring and Temporal-Decay Gradient Aggregation}
\label{sec:scale_2_inter}
To mitigate unbounded error cascades during long-term temporal adaptation, DISRL introduces an inter-window consolidation mechanism driven by first-order transition meta-gradients. The master parameter $\Theta_m$ provides a stable initialization anchor for fast policies across temporal windows. 
After intra-window adaptation,
the transferred policy is evaluated upon the arrival of
snapshot $t+1$, using the available training supervision
$\mathcal{C}_{\mathrm{rel}}^{t+1}$ to compute a transition meta-loss:


\begin{equation}
\resizebox{0.90\columnwidth}{!}{$\displaystyle
\mathcal{L}_{t\rightarrow t+1}^{\mathrm{meta}}(\Phi_{m,t+1})=-\sum_{i=1}^{B}\sum_{k=0}^{T_i-1}R_{i,k}^{t+1}\log\pi_{\Phi_{m,t+1}}\left(a_{i,k}^{t+1}\mid\mathbf{S}_{i,k}^{t+1}\right),
$}
\label{eq:transition_meta_loss}
\end{equation}
where $B$ denotes the meta-batch size, and $T_i$ denotes the exploration horizon of the $i$-th trajectory rollout. 

Deriving the exact meta-gradient with respect to the master parameter $\Theta_m$ requires backpropagating through the complete unrolled intra-window adaptation path: 
\begin{equation}
\nabla_{\Theta_m}
\mathcal{L}_{t\rightarrow t+1}^{\mathrm{meta}}
=
\left(
\frac{\partial \Phi_{m,t+1}}
{\partial \Theta_m}
\right)^{\top}
\nabla_{\Phi_{m,t+1}}
\mathcal{L}_{t\rightarrow t+1}^{\mathrm{meta}}.
\label{eq:exact_meta_gradient}
\end{equation}
Instead of computing the exact meta-gradient through the entire adaptation trajectory, DISRL adopts first-order meta-learning approximation~\cite{nichol-2018-firstorder} and directly estimates the transition correction:
\begin{equation}
\mathbf{g}_{t\rightarrow t+1}
=
\nabla_{\Phi_{m,t+1}}
\mathcal{L}_{t\rightarrow t+1}^{\mathrm{meta}}
(\Phi_{m,t+1}).
\label{eq:first_order_transition_gradient}
\end{equation}

These transition gradients capture the discrepancy between
current policy adaptation and future topology evolution
and are buffered without immediately
updating the fast policy.

At the end of each temporal window, DISRL aggregates transition gradients with temporal decay weights $\omega_t^m$ controlled by a decay factor $\rho \in (0,1]$:

\begin{align}
\omega_t^m
&=
\frac{\rho^{t_e-1-t}}
{\sum_{q=t_s}^{t_e-1}\rho^{t_e-1-q}},
\quad t=t_s,\ldots,t_e-1,
\label{eq:temporal_decay_weight}
\\
\mathbf{g}_m
&=
\sum_{t=t_s}^{t_e-1}
\omega_t^m\mathbf{g}_{t\rightarrow t+1},
\quad
\Theta_{m+1}=\Theta_m-\eta\mathbf{g}_m ,
\label{eq:master_update}
\end{align}
where recent transitions receive larger weights. The updated master parameter initializes the next window, providing stable policy evolution and reducing long-term tracking drift.

\subsection{Theoretical Analysis}
\label{sec:theoretical_analysis}

We theoretically analyze why continuous fast-parameter
transfer may suffer from long-horizon error accumulation
and how the proposed dual-scale mechanism mitigates
this issue. Detailed proofs are provided in
Appendix~\ref{sec:supp_theory}.
Let $\Phi_{m,t}$ and $\Phi_{m,t}^{\star}$ denote the
actual and ideal fast parameters at snapshot $G^t$,
respectively. We define the tracking error as
$d_{m,t}=\|\Phi_{m,t}-\Phi_{m,t}^{\star}\|_2$.
We consider three standard assumptions:
(1) bounded local adaptation perturbation
$\|\boldsymbol{\xi}_{m,t}\|_2\leq\xi$,
(2) bounded temporal drift
$\|\Phi_{m,t+1}^{\star}-\Phi_{m,t}^{\star}\|_2
\leq\Delta$, and
(3) a Lipschitz adaptation operator with coefficient
$\kappa$~\cite{hardt-2016-train}, which is expressed as
\begin{equation}
\|\mathcal{U}_t(\Phi_a)-\mathcal{U}_t(\Phi_b)\|_2
\leq\kappa\|\Phi_a-\Phi_b\|_2.
\label{eq:lipschitz_fast_adaptation}
\end{equation}

The perturbed fast adaptation is
\begin{equation}
\Phi_{m,t+1}=\Phi'_{m,t}
=\mathcal{U}_t(\Phi_{m,t})+\boldsymbol{\xi}_{m,t},
\label{eq:perturbed_fast_adaptation}
\end{equation}
where the ideal parameter satisfies the fixed-point
condition
$\mathcal{U}_t(\Phi_{m,t}^{\star})
=\Phi_{m,t}^{\star}$.
We establish the following proposition.

\noindent\textbf{Proposition 1 (Single-step Error
Propagation).}
For two consecutive snapshots within the same temporal
window, the fast-parameter tracking error satisfies:
\begin{equation}
d_{m,t+1}\leq\kappa d_{m,t}+\xi+\Delta.
\end{equation}

The proposition indicates that each snapshot transition
may introduce adaptation perturbations
and temporal drift, potentially contributing to error
accumulation during continuous policy transfer.
By recursively applying Proposition 1 within a temporal
window $\mathcal{W}_m$, the
tracking error satisfies the following
upper bound:
\begin{equation}
d_{m,t}\leq
\kappa^{t-t_s}d_{m,t_s}
+(\xi+\Delta)
\sum_{r=0}^{t-t_s-1}\kappa^r.
\end{equation}

Therefore, without periodic policy anchoring,
sequential fast-parameter transfer over long temporal
horizons may allow historical errors
to accumulate or amplify, increasing the risk of
error cascades, particularly when $\kappa \geq 1$.

\noindent\textbf{Corollary 1 (Window-confined Error
Propagation).}
In DISRL, the fast parameter is reinitialized
from the master parameter $\Theta_m$ at each
temporal window boundary.
Therefore, direct consecutive fast-parameter
transfer is restricted to at most $w-1$
transitions within a window of size
$w=|\mathcal{W}_m|$, yielding:

\begin{equation}
d_{m,t_e}\leq
d_{m,t_s}+(w-1)(\xi+\Delta),
\quad \text{when }\kappa=1.
\end{equation}

{
Hence, the dual-scale design restricts direct error
propagation within each temporal window, mitigating
the risk of long-horizon error accumulation.
This effect is further supported by the ablation
study in Section~4.3 and the long-horizon evaluation
in Section~\ref{sec:tracking_stability}.
}

\newif\ifDISRLCompactFigures
\DISRLCompactFigurestrue

\newcommand{\DISRLFigureScale}{1.00}
\newcommand{\DISRLFigureTopAdjust}{-8pt}
\newcommand{\DISRLFigureCaptionAdjust}{-7pt}
\newcommand{\DISRLFigureBottomAdjust}{-7pt}

\newcommand{\DISRLFigTwoTrim}{0pt 0pt 0pt 0pt}
\newcommand{\DISRLFigThreeTrim}{0pt 0pt 0pt 0pt}
\newcommand{\DISRLFigFourTrim}{0pt 6pt 0pt 0pt}
\newcommand{\DISRLFigFiveTrim}{0pt 3pt 0pt 3pt}

\newcommand{\DISRLFigureImage}[2]{%
  \ifDISRLCompactFigures
    \vspace*{\DISRLFigureTopAdjust}%
    \edef\DISRLExpandedImage{%
      \noexpand\includegraphics[width=\DISRLFigureScale\noexpand\columnwidth,trim=#1,clip]{#2}%
    }%
    \DISRLExpandedImage%
    \par\vspace{\DISRLFigureCaptionAdjust}%
  \else
    \includegraphics[width=\columnwidth]{#2}%
  \fi
}
\newcommand{\DISRLFigureEnd}{%
  \ifDISRLCompactFigures
    \vspace{\DISRLFigureBottomAdjust}%
  \fi
}

\section{Experiments}

\begin{table}[t]
\centering
\caption{Dataset statistics. $T$ denotes the number of snapshots;
$\overline{|\mathcal{V}|}$, $\overline{|\mathcal{E}|}$,
$\overline{|\mathcal{C}|}$, and $\overline{Q}$ denote the average
numbers of nodes, edges, and communities, and the average
modularity~\cite{newman-2006-modularity}, respectively.
All averages are taken over snapshots.}
\label{tab:dataset_statistics}
\begingroup
\normalsize
\setlength{\tabcolsep}{2.8pt}
\renewcommand{\arraystretch}{1.00}

\begin{tabular*}{\columnwidth}
{@{\extracolsep{\fill}}lrrrrr@{}}
\toprule
\textbf{Dataset}
& \multicolumn{1}{c}{\textbf{$T$}}
& \multicolumn{1}{c}{\textbf{$\overline{|\mathcal{V}|}$}}
& \multicolumn{1}{c}{\textbf{$\overline{|\mathcal{E}|}$}}
& \multicolumn{1}{c}{\textbf{$\overline{|\mathcal{C}|}$}}
& \multicolumn{1}{c}{\textbf{$\overline{Q}$}} \\
\midrule

\rowcolor{TableGroup}
\multicolumn{6}{@{}l}{\textit{Real-world dynamic dataset}} \\
Football & 24 & 137 & 653.9 & 19.8 & 0.5867 \\
\midrule

\rowcolor{TableGroup}
\multicolumn{6}{@{}l}{\textit{Expansion \& shrinkage setting}} \\
Amazon  & 25 & 18256.2  & 56573.5   & 1291.0 & 0.9301 \\
DBLP    & 25 & 132172.3 & 540405.8  & 4520   & 0.9105 \\
Twitter & 13 & 99566.3  & 1334284.8 & 2651   & 0.6313 \\
YouTube & 13 & 227400.2 & 1426123.3 & 2821.9 & 0.5422 \\
\midrule

\rowcolor{TableGroup}
\multicolumn{6}{@{}l}{\textit{Merging \& splitting setting}} \\
Amazon & 9 & 13178  & 38359.2  & 809.3  & 0.9837 \\
DBLP   & 9 & 114095 & 485261.7 & 2829.7 & 0.9164 \\
\bottomrule
\end{tabular*}

\endgroup
\end{table}

\begin{table*}[!t]
\centering
\caption{Main comparison on simulated evolutionary settings and a real-world dynamic dataset. {All metrics are reported in percentages (\%).} Best and second-best results are marked in bold and underlined within each comparison group, respectively.}
\label{tab:main_results}

\begingroup
\fontsize{8pt}{10pt}\selectfont
\setlength{\tabcolsep}{1.2pt}
\setlength{\aboverulesep}{0.25ex}
\setlength{\belowrulesep}{0.25ex}
\setlength{\cmidrulesep}{0.15ex}
\renewcommand{\arraystretch}{0.88}

\newcommand{\stdv}[2]{\mbox{#1\,$\pm$\,#2}}

\begin{tabular*}{\textwidth}{@{\extracolsep{\fill}}llcccc lcccc@{}}
\toprule
\textbf{Dataset}
& \textbf{Static}
& \textbf{Prec.}
& \textbf{Rec.}
& \textbf{F1}
& \textbf{Jac.}
& \textbf{Dynamic}
& \textbf{Prec.}
& \textbf{Rec.}
& \textbf{F1}
& \textbf{Jac.} \\
\midrule

\rowcolor{TableGroup}
\multicolumn{11}{@{}l}{
\textit{Real-world dynamic dataset}
} \\

\multirow{5}{*}{Football}
& \textbf{DISRL}
& \textbf{\boldmath\stdv{97.00}{0.21}}
& \textbf{\boldmath\stdv{94.32}{0.63}}
& \textbf{\boldmath\stdv{95.24}{0.36}}
& \textbf{\boldmath\stdv{91.33}{0.62}}
& \textbf{DISRL}
& \textbf{\boldmath\stdv{97.00}{0.21}}
& \stdv{94.32}{0.63}
& \textbf{\boldmath\stdv{95.24}{0.36}}
& \textbf{\boldmath\stdv{91.33}{0.62}} \\

& SLRL
& \uline{\stdv{91.99}{0.29}}
& \stdv{88.55}{0.90}
& \uline{\stdv{87.17}{0.67}}
& \uline{\stdv{81.03}{0.88}}
& CS-TGN
& \uline{\stdv{91.02}{4.27}}
& \textbf{\boldmath\stdv{96.22}{0.85}}
& \uline{\stdv{92.35}{1.63}}
& \uline{\stdv{87.53}{3.40}} \\

& CLARE
& \stdv{38.86}{1.10}
& \stdv{47.49}{1.39}
& \stdv{42.21}{1.19}
& \stdv{32.02}{1.14}
& MFC-TopoReg
& \stdv{65.38}{3.40}
& \stdv{62.18}{3.20}
& \stdv{63.42}{3.28}
& \stdv{52.92}{3.74} \\

& ProCom
& \stdv{29.90}{1.58}
& \stdv{54.08}{3.00}
& \stdv{34.41}{1.63}
& \stdv{23.58}{1.42}
& DevDynaP
& 88.11
& \uline{96.14}
& 91.80
& 86.17 \\

& SLSS
& 82.09
& \uline{90.00}
& 85.51
& 77.42
& QTCS
& 89.59
& 84.51
& 86.72
& 83.08 \\

\midrule

\rowcolor{TableGroup}
\multicolumn{11}{@{}l}{
\textit{Expansion\&shrinkage setting}
} \\

\multirow{5}{*}{Amazon}
& \textbf{DISRL}
& \textbf{\boldmath\stdv{86.78}{0.53}}
& \uline{\stdv{95.15}{0.43}}
& \textbf{\boldmath\stdv{89.86}{0.10}}
& \textbf{\boldmath\stdv{83.10}{0.16}}
& \textbf{DISRL}
& \uline{\stdv{86.78}{0.53}}
& \uline{\stdv{95.15}{0.43}}
& \textbf{\boldmath\stdv{89.86}{0.10}}
& \textbf{\boldmath\stdv{83.10}{0.16}} \\

& SLRL
& \stdv{79.77}{0.46}
& \textbf{\boldmath\stdv{97.34}{0.61}}
& \uline{\stdv{86.16}{0.66}}
& \uline{\stdv{77.59}{0.99}}
& CS-TGN
& \stdv{86.76}{0.69}
& \stdv{92.87}{0.72}
& \uline{\stdv{88.22}{0.02}}
& \stdv{80.68}{0.10} \\

& CLARE
& \stdv{86.06}{0.03}
& \stdv{75.36}{0.51}
& \stdv{78.54}{0.33}
& \stdv{68.45}{0.30}
& MFC-TopoReg
& \stdv{78.23}{0.08}
& \stdv{93.75}{0.09}
& \stdv{84.10}{0.08}
& \stdv{74.81}{0.10} \\

& ProCom
& \uline{\stdv{86.07}{0.28}}
& \stdv{75.79}{0.56}
& \stdv{78.49}{0.45}
& \stdv{69.48}{0.61}
& DevDynaP
& 82.73
& \textbf{96.03}
& 87.85
& 80.01 \\

& SLSS
& 81.13
& 93.60
& 85.36
& 77.38
& QTCS
& \textbf{93.44}
& 86.79
& 88.01
& \uline{81.38} \\

\midrule

\multirow{5}{*}{DBLP}
& \textbf{DISRL}
& \textbf{\boldmath\stdv{76.10}{0.51}}
& \uline{\stdv{83.83}{0.49}}
& \textbf{\boldmath\stdv{78.48}{0.46}}
& \textbf{\boldmath\stdv{70.26}{0.38}}
& \textbf{DISRL}
& \uline{\stdv{76.10}{0.51}}
& \textbf{\boldmath\stdv{83.83}{0.49}}
& \textbf{\boldmath\stdv{78.48}{0.46}}
& \textbf{\boldmath\stdv{70.26}{0.38}} \\

& SLRL
& \stdv{56.92}{1.30}
& \textbf{\boldmath\stdv{90.90}{0.57}}
& \uline{\stdv{68.47}{0.40}}
& \stdv{55.09}{0.51}
& CS-TGN
& \textbf{\boldmath\stdv{76.31}{0.47}}
& \stdv{82.97}{2.83}
& \stdv{75.63}{0.65}
& \stdv{66.12}{0.76} \\

& CLARE
& \stdv{60.85}{0.79}
& \stdv{61.83}{0.34}
& \stdv{59.98}{0.49}
& \stdv{49.88}{0.62}
& MFC-TopoReg
& \stdv{53.51}{0.24}
& \stdv{72.89}{0.23}
& \stdv{60.85}{0.24}
& \stdv{49.24}{0.22} \\

& ProCom
& \uline{\stdv{64.60}{0.73}}
& \stdv{72.02}{1.31}
& \stdv{66.20}{1.03}
& \uline{\stdv{58.16}{0.99}}
& DevDynaP
& 61.96
& \uline{83.77}
& 69.57
& 58.46 \\

& SLSS
& 56.71
& 81.62
& 65.38
& 54.27
& QTCS
& 74.96
& 81.92
& \uline{76.63}
& \uline{69.25} \\

\midrule

\multirow{5}{*}{YouTube}
& \textbf{DISRL}
& \textbf{\boldmath\stdv{72.02}{0.31}}
& \uline{\stdv{70.93}{0.32}}
& \textbf{\boldmath\stdv{67.76}{0.03}}
& \textbf{\boldmath\stdv{56.87}{0.08}}
& \textbf{DISRL}
& \textbf{\boldmath\stdv{72.02}{0.31}}
& \uline{\stdv{70.93}{0.32}}
& \textbf{\boldmath\stdv{67.76}{0.03}}
& \textbf{\boldmath\stdv{56.87}{0.08}} \\

& SLRL
& \uline{\stdv{57.16}{1.04}}
& \textbf{\boldmath\stdv{74.76}{1.63}}
& \uline{\stdv{57.52}{1.74}}
& \stdv{44.21}{1.50}
& CS-TGN
& \uline{\stdv{60.02}{1.00}}
& \textbf{\boldmath\stdv{80.28}{1.06}}
& \uline{\stdv{64.57}{0.49}}
& \stdv{51.77}{0.56} \\

& CLARE
& \stdv{51.93}{0.22}
& \stdv{42.74}{0.27}
& \stdv{44.72}{0.25}
& \stdv{33.94}{0.34}
& MFC-TopoReg
& \stdv{29.49}{0.55}
& \stdv{38.85}{0.59}
& \stdv{32.14}{0.54}
& \stdv{22.23}{0.42} \\

& ProCom
& \stdv{52.23}{0.39}
& \stdv{49.94}{0.93}
& \stdv{48.81}{0.78}
& \stdv{38.97}{0.70}
& DevDynaP
& 50.74
& 61.99
& 53.67
& 41.72 \\

& SLSS
& 53.24
& 63.66
& 56.01
& \uline{45.48}
& QTCS
& 58.60
& 66.00
& 60.08
& \uline{51.90} \\

\midrule

\multirow{5}{*}{Twitter}
& \textbf{DISRL}
& \textbf{\boldmath\stdv{65.32}{1.39}}
& \uline{\stdv{64.11}{0.68}}
& \textbf{\boldmath\stdv{61.30}{0.41}}
& \textbf{\boldmath\stdv{48.17}{0.53}}
& \textbf{DISRL}
& \textbf{\boldmath\stdv{65.32}{1.39}}
& \uline{\stdv{64.11}{0.68}}
& \textbf{\boldmath\stdv{61.30}{0.41}}
& \textbf{\boldmath\stdv{48.17}{0.53}} \\

& SLRL
& \uline{\stdv{48.35}{0.06}}
& \textbf{\boldmath\stdv{77.34}{0.23}}
& \uline{\stdv{56.44}{0.02}}
& \uline{\stdv{41.57}{0.07}}
& CS-TGN
& \uline{\stdv{49.16}{2.84}}
& \textbf{\boldmath\stdv{88.49}{1.28}}
& \uline{\stdv{57.05}{1.81}}
& \uline{\stdv{43.56}{1.93}} \\

& CLARE
& \stdv{39.24}{0.59}
& \stdv{35.15}{0.48}
& \stdv{35.30}{0.52}
& \stdv{23.66}{0.47}
& MFC-TopoReg
& \stdv{31.81}{0.81}
& \stdv{40.57}{1.17}
& \stdv{34.41}{0.92}
& \stdv{23.92}{0.69} \\

& ProCom
& \stdv{46.85}{0.84}
& \stdv{50.32}{1.11}
& \stdv{42.63}{0.61}
& \stdv{31.55}{0.63}
& DevDynaP
& 33.87
& 56.80
& 40.87
& 28.85 \\

& SLSS
& 47.44
& 59.72
& 50.88
& 38.34
& QTCS
& 38.33
& 53.27
& 42.83
& 30.92 \\

\midrule

\rowcolor{TableGroup}
\multicolumn{11}{@{}l}{
\textit{Merging\&splitting setting}
} \\

\multirow{5}{*}{Amazon}
& \textbf{DISRL}
& \stdv{77.71}{0.57}
& \uline{\stdv{95.51}{0.28}}
& \textbf{\boldmath\stdv{83.49}{0.50}}
& \textbf{\boldmath\stdv{73.97}{0.75}}
& \textbf{DISRL}
& \stdv{77.71}{0.57}
& \uline{\stdv{95.51}{0.28}}
& \textbf{\boldmath\stdv{83.49}{0.50}}
& \textbf{\boldmath\stdv{73.97}{0.75}} \\

& SLRL
& \stdv{68.89}{1.08}
& \textbf{\boldmath\stdv{96.84}{0.88}}
& \uline{\stdv{77.00}{1.35}}
& \uline{\stdv{66.36}{1.97}}
& CS-TGN
& \uline{\stdv{84.32}{0.24}}
& \stdv{75.38}{1.35}
& \stdv{76.32}{1.03}
& \stdv{64.69}{1.18} \\

& CLARE
& \textbf{\boldmath\stdv{86.03}{0.24}}
& \stdv{63.85}{0.42}
& \stdv{69.12}{0.36}
& \stdv{56.65}{0.38}
& MFC-TopoReg
& \stdv{71.24}{0.18}
& \textbf{\boldmath\stdv{96.43}{0.20}}
& \uline{\stdv{80.00}{0.19}}
& \uline{\stdv{69.10}{0.21}} \\

& ProCom
& \uline{\stdv{85.72}{0.30}}
& \stdv{62.80}{0.66}
& \stdv{68.05}{0.40}
& \stdv{56.05}{0.54}
& DevDynaP
& 79.64
& 67.00
& 66.71
& 55.37 \\

& SLSS
& 84.05
& 73.69
& 73.70
& 62.19
& QTCS
& \textbf{95.91}
& 57.69
& 66.95
& 55.65 \\

\midrule

\multirow{5}{*}{DBLP}
& \textbf{DISRL}
& \uline{\stdv{53.34}{1.02}}
& \uline{\stdv{75.09}{1.40}}
& \textbf{\boldmath\stdv{56.63}{1.27}}
& \textbf{\boldmath\stdv{44.56}{1.49}}
& \textbf{DISRL}
& \uline{\stdv{53.34}{1.02}}
& \textbf{\boldmath\stdv{75.09}{1.40}}
& \textbf{\boldmath\stdv{56.63}{1.27}}
& \textbf{\boldmath\stdv{44.56}{1.49}} \\

& SLRL
& \stdv{37.76}{1.05}
& \textbf{\boldmath\stdv{86.67}{0.19}}
& \stdv{46.74}{0.14}
& \stdv{34.63}{0.17}
& CS-TGN
& \stdv{48.24}{2.19}
& \uline{\stdv{68.91}{2.14}}
& \uline{\stdv{50.21}{0.87}}
& \stdv{36.51}{0.86} \\

& CLARE
& \stdv{53.28}{0.93}
& \stdv{49.05}{0.53}
& \stdv{45.39}{0.09}
& \stdv{33.57}{0.12}
& MFC-TopoReg
& \stdv{25.64}{0.28}
& \stdv{57.03}{0.31}
& \stdv{31.74}{0.31}
& \stdv{21.66}{0.25} \\

& ProCom
& \textbf{\boldmath\stdv{55.76}{1.00}}
& \stdv{55.60}{0.76}
& \uline{\stdv{49.92}{0.54}}
& \uline{\stdv{38.04}{0.47}}
& DevDynaP
& 50.95
& 53.93
& 45.39
& 33.90 \\

& SLSS
& 44.50
& 56.26
& 44.75
& 33.61
& QTCS
& \textbf{63.40}
& 49.80
& 49.61
& \uline{39.39} \\

\bottomrule
\end{tabular*}
\endgroup
\end{table*}

\subsection{Experimental Setup}
\textbf{Dynamic Benchmark Construction.} Existing benchmarks provide limited support for 
dynamic local community tracking: real-world datasets are mostly static, while synthetic evolutionary benchmarks (e.g., CoDÆN~\cite{paoletti-2025-codaen}) mainly focus on global community partitioning and fail to simulate sparse, seed-centered local tracking. To address this limitation, we construct a non-stationary benchmark by synthesizing controlled, chronologically ordered evolutionary mutations onto four empirical networks while preserving their original topological complexity and community fidelity.  
Specifically, we generate temporal graph streams through four representative evolutionary operations, including community expansion, shrinkage, merging, and splitting, together with background topological perturbations (random node/edge insertions and deletions). The ground-truth communities are synchronously updated with graph evolution to provide reliable supervision.

\noindent\textbf{Datasets and Evaluation Scenarios.} We evaluate DISRL on five datasets covering simulated evolution and real-world dynamics. For real-world dynamics, we use Football, derived from the Century of College Football dataset~\cite{century-college-football-dataset}, comprising 24 yearly snapshots from 2000--2024, with real game interactions and conference-based community labels. For simulated evolution, we use four SNAP datasets\footnote{\url{https://snap.stanford.edu/data/}} with authentic community annotations: Amazon, DBLP, YouTube, and Twitter. We consider two scenarios: (1) expansion\&shrinkage on all four datasets, which evaluates gradual community growth and contraction; and (2) merging\&splitting on Amazon and DBLP, which tests structural reorganization among communities. Dataset statistics are summarized in Table~\ref{tab:dataset_statistics}. Detailed generation procedures and evaluation
protocols are provided in Appendix~\ref{sec:supp_evaluation_settings}.

\noindent\textbf{Baselines.} We compare DISRL with representative baselines from two categories: 

 \textit{(i) Static Local Baselines:} Applied independently to isolated snapshots using snapshot-specific topologies and annotations. \textbf{SLSS}~\cite{ni-2024-semisupervised} leverages graph-kernel similarity for local expansion; \textbf{SLRL}~\cite{ni-2025-slrl} uses an RL-based extractor-expander mechanism; \textbf{CLARE}~\cite{wu-2022-clare} operates on a locator-rewriter framework; and \textbf{ProCom}~\cite{wu-2024-procom} adopts a pre-training prompt paradigm for few-shot target localization.
 
\textit{(ii) Dynamic Tracking Baselines:} Optimized across temporal streams. \textbf{DevDynaP}~\cite{bakhtar-2025-devdynap} executes incremental heuristic exploration; \textbf{MFC-TopoReg}~\cite{kong-2024-learning} enforces topological regularization over dynamic matrix factorization partitions; \textbf{QTCS}~\cite{lin-2024-qtcs} optimizes query-centered temporal searches; and \textbf{CS-TGN}~\cite{hashemi-2023-cstgn} employs a query-driven temporal GNN.

\noindent\textbf{Metrics and Implementation Details.}
Following~\cite{ni-2024-semisupervised,ni-2025-slrl},
we evaluate tracking accuracy across snapshots by comparing
predictions against evolving ground-truth boundaries using four
metrics: Precision, Recall, F1-score, and Jaccard.
All metrics are computed per seed-snapshot pair
and macro-averaged across all evaluated seeds and snapshots.

DISRL is implemented in PyTorch on an NVIDIA RTX 3090 GPU (24GB).
For fair comparison, \textit{MFC-TopoReg} is post-processed to
extract the community containing the target seed;
query-based methods (\textit{CS-TGN}, \textit{QTCS}, and
\textit{DevDynaP}) are evaluated under single-seed queries.
Following~\cite{ni-2025-slrl}, the number of generated candidate
communities for CLARE and ProCom is set to $5000$.
Stochastic methods are evaluated over five random seeds,
with results reported as mean $\pm$ standard deviation;
deterministic methods are evaluated once. {
For each target seed, we exclude its ground-truth community
from the available supervision. Specifically, Football uses
all remaining annotated communities, while the other datasets
randomly sample 100 communities per snapshot and remove the
target community if present. Each target seed is tracked
independently with its own policy parameters. Snapshots are
processed chronologically, using only current and historical
supervision, while target-community labels are reserved
for evaluation.
}
The detailed hyper-parameter configurations of DISRL are
provided in Appendix Table~\ref{tab:disrl_hyperparameters}.
We highlight three key parameters corresponding to our proposed
modules:
the number of retrieved relevant communities
$K=5$ for Football and $K=50$ for the other datasets,
iGPN propagation layers
$L=2$ for Football and $L=3$ for the other datasets,
and temporal window size $w=4$ for all datasets. {The anonymized source code is available at
\url{https://anonymous.4open.science/r/DISRL-code-5D02}.}

\begin{table*}[!t]
\centering
\caption{Ablation results. Each cell reports F1/Jaccard.}
\label{tab:ablation_results}

\begingroup
\fontsize{7.8pt}{9pt}\selectfont
\setlength{\tabcolsep}{0.5pt}
\setlength{\aboverulesep}{0.20ex}
\setlength{\belowrulesep}{0.20ex}
\setlength{\cmidrulesep}{0.10ex}
\renewcommand{\arraystretch}{1.00}

\newcommand{\stdv}[2]{%
\mbox{#1\,$\pm$\,#2}%
}

\newcommand{\fj}[2]{%
\mbox{#1\hspace{0.1pt}/\hspace{0.1pt}#2}%
}

\begin{tabular*}{\textwidth}
{@{\extracolsep{\fill}}lccccc@{}}
\toprule

\multirow{2}{*}{\textbf{Variant}}
& \multicolumn{1}{c}{\textbf{Real-world dynamics}}
& \multicolumn{2}{c}{\textbf{Expansion\&shrinkage}}
& \multicolumn{2}{c}{\textbf{Merging\&splitting}} \\

\cmidrule(lr){2-2}
\cmidrule(lr){3-4}
\cmidrule(lr){5-6}

& \textbf{Football}
& \textbf{Amazon}
& \textbf{DBLP}
& \textbf{Amazon}
& \textbf{DBLP} \\

\midrule

w/o cc
&
\fj
{\stdv{93.42}{0.27}}
{\stdv{87.94}{0.44}}
&
\fj
{\stdv{88.02}{0.12}}
{\stdv{80.32}{0.16}}
&
\fj
{\stdv{75.89}{0.13}}
{\stdv{67.39}{0.23}}
&
\fj
{\stdv{81.48}{0.52}}
{\stdv{71.40}{0.87}}
&
\fj
{\stdv{52.54}{1.18}}
{\stdv{40.81}{1.40}}
\\

w/o rcs

&
\fj
{\stdv{94.23}{0.30}}
{\stdv{90.18}{0.49}}
&
\fj
{\stdv{88.28}{0.15}}
{\stdv{80.63}{0.19}}
&
\fj
{\stdv{76.53}{0.24}}
{\stdv{67.67}{0.32}}
&
\fj
{\stdv{82.86}{0.44}}
{\stdv{73.35}{0.31}}
&
\fj
{\stdv{53.58}{1.08}}
{\stdv{41.65}{1.24}}

\\

w/o sim
&
\fj
{\stdv{88.29}{1.86}}
{\stdv{81.95}{2.15}}
&
\fj
{\stdv{87.28}{1.00}}
{\stdv{79.27}{1.42}}
&
\fj
{\stdv{75.29}{1.71}}
{\stdv{66.80}{1.66}}
&
\fj
{\stdv{81.76}{1.94}}
{\stdv{71.46}{1.95}}
&
\fj
{\stdv{53.25}{2.24}}
{\stdv{41.72}{2.49}}
\\

w/o win
&
\fj
{\stdv{92.97}{0.46}}
{\stdv{88.20}{0.62}}
&
\fj
{\stdv{87.14}{0.54}}
{\stdv{78.92}{0.64}}
&
\fj
{\stdv{74.81}{0.41}}
{\stdv{65.07}{0.76}}
&
\fj
{\stdv{79.96}{1.14}}
{\stdv{69.68}{1.17}}
&
\fj
{\stdv{49.07}{1.65}}
{\stdv{37.66}{1.53}}
\\



w/o decay
&
\fj
{\stdv{94.86}{0.30}}
{\stdv{90.67}{0.49}}
&
\fj
{\stdv{88.53}{0.46}}
{\stdv{81.17}{0.80}}
&
\fj
{\stdv{77.08}{0.84}}
{\stdv{68.57}{1.12}}
&
\fj
{\stdv{82.46}{0.36}}
{\stdv{72.84}{0.54}}
&
\fj
{\stdv{54.98}{1.58}}
{\stdv{43.20}{1.40}}
\\

\textbf{DISRL}
&
\fj
{\textbf{\boldmath\stdv{95.24}{0.36}}}
{\textbf{\boldmath\stdv{91.33}{0.62}}}
&
\fj
{\textbf{\boldmath\stdv{89.86}{0.10}}}
{\textbf{\boldmath\stdv{83.10}{0.16}}}
&
\fj
{\textbf{\boldmath\stdv{78.48}{0.46}}}
{\textbf{\boldmath\stdv{70.26}{0.38}}}
&
\fj
{\textbf{\boldmath\stdv{83.49}{0.50}}}
{\textbf{\boldmath\stdv{73.97}{0.75}}}
&
\fj
{\textbf{\boldmath\stdv{56.63}{1.27}}}
{\textbf{\boldmath\stdv{44.56}{1.49}}}
\\

\bottomrule
\end{tabular*}

\endgroup
\end{table*}

\subsection{Main Results}

\noindent\textbf{Comparison with Static Baselines.}
As shown in Table~\ref{tab:main_results}, DISRL achieves the best $\text{F1-score}$ and $\text{Jaccard similarity}$ on the real-world Football dataset and across all four benchmarks under expansion\&shrinkage dynamics. Compared with the strongest static baseline, DISRL obtains absolute $\text{F1}$ gains of {8.07, 3.70, 10.01, 10.24, and 4.86 percentage points (pp)} on Football, Amazon, DBLP, YouTube, and Twitter, respectively. Static baselines suffer from \textit{temporal isolation}: processing snapshots independently prevents them from using historical trajectories to alleviate sparse supervision. Candidate-generation methods such as \textit{CLARE} and \textit{ProCom} further optimize global candidate pools rather than seed-centered local targets, often producing mismatched boundaries. In contrast, DISRL combines active seed-centric MDP rollouts with adaptive supervision construction, enabling temporally consistent tracking. Under merging\&splitting dynamics, DISRL also maintains top $\text{F1}$ and $\text{Jaccard}$ performance on both Amazon and DBLP, showing robustness to abrupt boundary mutations.

\noindent\textbf{Comparison with Dynamic Baselines.}
Table~\ref{tab:main_results} compares DISRL with dynamic tracking models. DISRL achieves the best $\text{F1-score}$ and $\text{Jaccard similarity}$ across all settings. {On the real-world Football dataset,
DISRL achieves an F1 gain of 2.89 pp
over the strongest dynamic baseline.
Under expansion\&shrinkage and merging\&splitting
settings, its largest F1 gains reach 4.25 pp
on Twitter and 6.42 pp on DBLP, respectively.} Although dynamic baselines exploit temporal information, they remain weakly aligned with seed-centered tracking. Heuristic methods such as \textit{DevDynaP} and \textit{QTCS} rely on topological propagation or proximity thresholds, making them sensitive to edge noise. Matrix-factorization methods such as \textit{MFC-TopoReg} prioritize global clustering consistency and may sacrifice local boundary fidelity. Temporal GNNs such as \textit{CS-TGN} capture structural evolution but still rely on passive node representations without explicit boundary expansion or stopping decisions. In contrast, DISRL dynamically constructs seed-specific supervision and employs dual-scale policy updating, effectively reducing supervision mismatch and error accumulation in evolving graphs.


\subsection{Ablation Study}

{To investigate the contributions of
the modules of DISRL, we conduct five ablation
experiments, with results reported in
Table~\ref{tab:ablation_results}.}

\noindent\textbf{{Supervision Construction Module.}}
{To evaluate reference community synthesis,
we consider two variants:
\textit{w/o cc}, which removes connected-component
extraction and directly inherits previous predictions
as reference communities; and
\textit{w/o rcs}, which replaces reference community
synthesis with the two-hop neighborhood of the target
seed. Their performance degradation highlights the
importance of constructing topology-consistent reference
communities for reliable seed-specific supervision.
To assess relevant community retrieval, we further
introduce \textit{w/o sim}, which randomly selects
$K$ known communities instead of using structural
similarity. Its inferior performance suggests that
structural similarity-based retrieval helps reduce
supervision mismatch and provides more relevant guidance
for community boundary expansion.}

\noindent\textbf{{Dual-scale Community Exploration Module.}}
{To investigate the effect of window-level
master anchoring, we consider}
\textit{w/o win}, which removes window-level master
anchoring and continuously transfers the fast policy
across snapshots.
{Its performance degradation suggests
that window-level anchoring helps mitigate the
accumulation of localized adaptation errors across
snapshots. We further evaluate temporal-decay gradient
aggregation through \textit{w/o decay}, which replaces
temporal-decay weighting with uniform averaging.
The inferior performance of this variant indicates
that emphasizing more recent transition gradients
benefits policy adaptation across windows.
Overall, DISRL consistently outperforms all five
variants under real-world dynamics,
expansion\&shrinkage, and merging\&splitting settings.}


\subsection{{Long-Horizon Error Mitigation Analysis}}
\label{sec:tracking_stability}

\begin{figure}[!htbp]
    \centering
    \DISRLFigureImage{\DISRLFigTwoTrim}{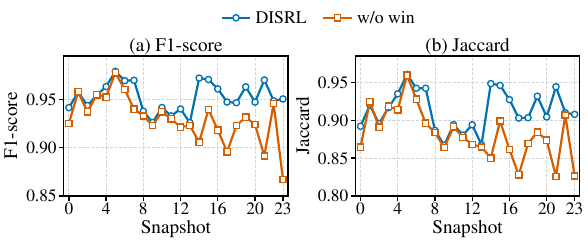}
    \caption{{Long-horizon tracking performance on Football.}}
    \label{fig:tracking_stability}
\DISRLFigureEnd
\end{figure}

{
To empirically investigate the error-mitigation
mechanism, we compare DISRL with its window-free
variant (\textit{w/o win}) across 24 consecutive
snapshots of the real-world Football dataset,
spanning six temporal windows ($w=4$).
As illustrated in Fig.~\ref{fig:tracking_stability},
both variants achieve comparable performance in the
early snapshots, whereas their performance diverges
more noticeably in later stages. Specifically, the
window-free variant exhibits more pronounced performance
degradation and fluctuations, while DISRL maintains
relatively high F1-score and Jaccard similarity
throughout the sequence.
As established by Corollary~1 in Section~3.4,
periodic master-policy anchoring restricts direct
fast-parameter error propagation to each temporal
window. These experimental observations are consistent
with this theoretical analysis, suggesting that the
dual-scale mechanism helps mitigate the adverse
effects of accumulated adaptation errors on
long-horizon tracking performance.
}

\subsection{{Hyper-parameter Sensitivity}}
\label{sec:additional_analysis}

\label{sec:supp_hyperparameter_sensitivity}

\begin{figure}[!htbp]
\centering
\DISRLFigureImage{\DISRLFigThreeTrim}{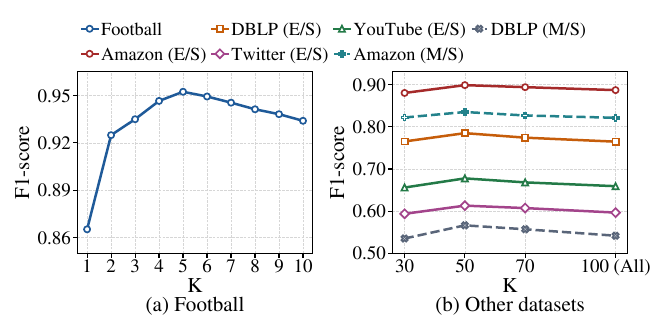}
\caption{Effect of the retrieved community number $K$.}
\Description{{F1 and Jaccard scores for different retrieval sizes on Football, Amazon, and DBLP.}}
\label{fig:topk_study}
\DISRLFigureEnd
\end{figure}

\noindent\textbf{Retrieved relevant communities.} We report the supervision retrieval parameter $K$ in Fig.~\ref{fig:topk_study}. For the simulated-evolution datasets, DISRL achieves stable performance with $K=50$, balancing supervision coverage and structural relevance. When $K=100$, all known communities are included, leading to performance degradation due to potentially irrelevant supervision. On the Football dataset, we evaluate $K$ from 1 to 10, with the best F1-score achieved at $K=5$. Performance initially improves as $K$ increases, but gradually declines beyond this point, highlighting the importance of selecting an appropriate number of relevant communities.

\begin{figure}[!htbp]
\centering
\DISRLFigureImage{\DISRLFigFourTrim}{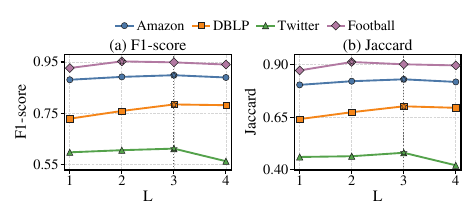}
\caption{Sensitivity to iGPN propagation.}
\Description{{F1 and Jaccard scores on Football, Amazon, DBLP, and Twitter for different propagation depths.}}
\label{fig:igpn_layers_study}
\DISRLFigureEnd
\end{figure}

\noindent\textbf{iGPN Propagation Layers.}\label{sec:supp_igpn_layers} We then study the sensitivity of DISRL to the number of iGPN propagation
layers $L$ used by the localized state encoder
on the real-world Football dataset and three
expansion\&shrinkage benchmarks (Amazon, DBLP, and Twitter).
As shown in
Fig.~\ref{fig:igpn_layers_study},
$L=2$ achieves the best performance on Football,
whereas $L=3$ achieves the best performance on Amazon, DBLP,
and Twitter.
As $L$ controls the depth of localized
structural propagation, too few layers may miss multi-hop boundary context,
whereas too many layers may introduce remote structural noise into
seed-centered representations.
These results suggest that two-layer propagation is sufficient
for Football, while three-layer localized propagation
better balances structural context modeling and noise control
on the other datasets.

\begin{figure}[!htbp]
\centering
\renewcommand{\DISRLFigureCaptionAdjust}{-9pt}
\renewcommand{\DISRLFigureBottomAdjust}{-9pt}
\DISRLFigureImage{\DISRLFigFiveTrim}{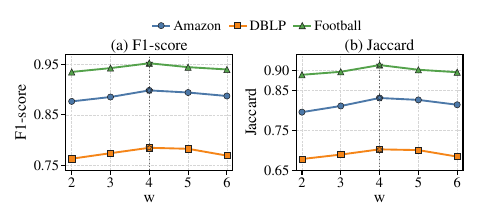}
\caption{Sensitivity to temporal window size.}
\Description{{F1 and Jaccard scores on Football, Amazon, and DBLP for window sizes from two to six.}}
\label{fig:window_size_study}
\DISRLFigureEnd
\end{figure}

\noindent\textbf{Temporal Window Horizon.}\label{sec:supp_window_size} We further evaluate the sensitivity of DISRL to the temporal
window size $w \in \{2, 3, 4, 5, 6\}$
on Football, Amazon, and DBLP, with the latter two evaluated under the expansion\&shrinkage setting.
As illustrated in Fig.~\ref{fig:window_size_study}, $w=4$
achieves the best overall performance across these three datasets,
indicating a trade-off between policy stabilization and temporal drift:
\begin{itemize}
    \item \textbf{Under-sized Windows ($w=2$):} Restricting the horizon
    induces \textit{temporal myopia}. The fast parameters ($\Phi_{m,t}$)
    overfit to transient single-frame fluctuations, starving the master
    anchor ($\Theta_m$) of continuous, long-horizon meta-gradients.
    \item \textbf{Over-sized Windows ($w=6$):} Expanding the horizon
    excessively introduces optimization lag. Under non-stationary graph
    streams, historical transition gradients from distant snapshots carry
    stale topological signals, causing \textit{gradient smearing} that
    misaligns the window-level master update.
\end{itemize}
Consequently, setting the default horizon to $w=4$ harmonizes short-term
micro-evolution tracking with long-term macro-trend anchoring,
mitigating the risk of long-horizon error accumulation.

\subsection{{Additional Experimental Analyses}}
\label{sec:additional_experiments}

{The qualitative case study and efficiency comparison are provided in Appendices~\ref{sec:supp_case_study} and~\ref{sec:supp_efficiency}, respectively. DISRL achieves up to $10.35\times$ speedup over CS-TGN and effectively tracks evolving community boundaries through adaptive supervision construction and dual-scale policy consolidation.}

\FloatBarrier

\section{Conclusion}
In this paper, we proposed DISRL, a dual-scale incremental semi-supervised reinforcement learning framework for dynamic local community tracking. DISRL addresses temporal supervision uncertainty by constructing seed-specific adaptive guidance from sparse communities, and mitigates long-horizon error cascades through a dual-scale policy optimization strategy. We further developed a dynamic graph benchmark with diverse community evolution patterns. Extensive experiments demonstrate that DISRL consistently outperforms state-of-the-art baselines with improved efficiency and tracking stability.

\begin{acks}
This work was supported by the National Natural Science Foundation
of China under Grant Nos. 62302451 and 62276233.
\end{acks}

\bibliographystyle{ACM-Reference-Format}
\bibliography{refer}

\appendix

\section{Theoretical Analysis}
\label{sec:supp_theory}

We analyze how localized adaptation errors propagate and accumulate during
sequential fast-parameter transfer within a temporal window
$\mathcal{W}_m$ spanning from $t_s$ to $t_e$. Let $\Phi_{m,t}$ and
$\Phi_{m,t}^{\star}$ denote the actual fast parameter and the ideal
error-free parameter at snapshot $G^t$, respectively.

Following time-varying optimization analyses~\cite{madden-2020-bounds}, the
snapshot-level tracking error is defined as
$d_{m,t}=\left\|\Phi_{m,t}-\Phi_{m,t}^{\star}\right\|_2$, using the
standard L2 norm. We establish the tracking dynamics under three standard
stability assumptions.

\noindent\textbf{Assumption 1 (Bounded Adaptation Perturbation).}
The local optimization perturbation $\boldsymbol{\xi}_{m,t}$ introduced
during adaptation on $G^t$ is uniformly bounded; that is, there exists a
constant $\xi\geq 0$ such that
\begin{equation}
\left\|\boldsymbol{\xi}_{m,t}\right\|_2\leq\xi.
\label{eq:supp_bounded_local_perturbation}
\end{equation}

\noindent\textbf{Assumption 2 (Bounded Ideal-parameter Drift).}
The temporal variation of the ideal tracking target across adjacent
snapshots is bounded; that is, there exists a constant $\Delta\geq 0$ such
that
\begin{equation}
\left\|\Phi_{m,t+1}^{\star}-\Phi_{m,t}^{\star}\right\|_2\leq\Delta.
\label{eq:supp_bounded_ideal_policy_drift}
\end{equation}

\noindent\textbf{Assumption 3 (Lipschitz Adaptation Operator).}
For theoretical analysis, let $\mathcal{U}_t(\cdot)$ denote the error-free
counterpart of the snapshot-level adaptation on $G^t$. Corresponding to the
policy-gradient update and fast-parameter transfer defined in the main
paper, the adapted parameter and its subsequent transfer can be written as
\begin{equation}
\Phi_{m,t}'
=\mathcal{U}_t(\Phi_{m,t})+\boldsymbol{\xi}_{m,t},
\qquad
\Phi_{m,t+1}\leftarrow\Phi_{m,t}'.
\label{eq:supp_perturbed_fast_adaptation}
\end{equation}
Following standard stability analyses of gradient-based
optimization~\cite{hardt-2016-train}, we assume that
$\mathcal{U}_t(\cdot)$ is uniformly Lipschitz continuous. Specifically,
there exists a constant $\kappa\geq 0$ such that
\begin{equation}
\left\|\mathcal{U}_t(\Phi_a)-\mathcal{U}_t(\Phi_b)\right\|_2
\leq
\kappa\left\|\Phi_a-\Phi_b\right\|_2.
\label{eq:supp_lipschitz_fast_adaptation}
\end{equation}
The ideal parameter $\Phi_{m,t}^{\star}$ is assumed to be a fixed point of
the error-free adaptation on $G^t$, i.e.,
$\mathcal{U}_t(\Phi_{m,t}^{\star})=\Phi_{m,t}^{\star}$.

\noindent\textbf{Proposition 1 (Single-step Error Propagation).}
Under Assumptions~1--3, for any adjacent snapshots $G^t$ and $G^{t+1}$
within the same temporal window $\mathcal{W}_m$, the fast-parameter error
satisfies
\begin{equation}
d_{m,t+1}\leq\kappa d_{m,t}+\xi+\Delta.
\label{eq:supp_single_step_error_recursion}
\end{equation}

\noindent\textit{Proof.}
By the definition of $d_{m,t+1}$ and the fast-parameter transfer relation in
Eq.~\eqref{eq:supp_perturbed_fast_adaptation}, we have
\begin{align}
d_{m,t+1}
&=\left\|\Phi_{m,t+1}-\Phi_{m,t+1}^{\star}\right\|_2
\nonumber\\
&=\left\|\mathcal{U}_t(\Phi_{m,t})+\boldsymbol{\xi}_{m,t}
-\Phi_{m,t+1}^{\star}\right\|_2 \nonumber\\
&\leq
\left\|\mathcal{U}_t(\Phi_{m,t})
-\mathcal{U}_t(\Phi_{m,t}^{\star})\right\|_2
+\left\|\boldsymbol{\xi}_{m,t}\right\|_2 \nonumber\\
&\quad+
\left\|\mathcal{U}_t(\Phi_{m,t}^{\star})
-\Phi_{m,t+1}^{\star}\right\|_2. \nonumber
\end{align}
The first term is bounded by $\kappa d_{m,t}$ by Assumption~3, and the
second term is bounded by $\xi$ by Assumption~1. For the third term, using
$\mathcal{U}_t(\Phi_{m,t}^{\star})=\Phi_{m,t}^{\star}$ and Assumption~2,
we obtain
\begin{equation*}
\begin{aligned}
&\left\|\mathcal{U}_t(\Phi_{m,t}^{\star})-\Phi_{m,t+1}^{\star}\right\|_2\\
&\quad=\left\|\Phi_{m,t}^{\star}-\Phi_{m,t+1}^{\star}\right\|_2
\leq\Delta.
\end{aligned}
\end{equation*}
Combining the three bounds yields
Eq.~\eqref{eq:supp_single_step_error_recursion}.
\hfill$\square$

\noindent\textbf{Long-horizon Error Accumulation.}
By recursively applying Eq.~\eqref{eq:supp_single_step_error_recursion}
from the window start $t_s$ to any snapshot $t\in\mathcal{W}_m$, we obtain
\begin{equation}
d_{m,t}
\leq
\kappa^{t-t_s}d_{m,t_s}
+(\xi+\Delta)
\sum_{r=0}^{t-t_s-1}\kappa^r.
\label{eq:supp_window_error_unrolling}
\end{equation}

When $0\leq\kappa<1$, the geometric sum is bounded by
$1/(1-\kappa)$, so the tracking-error upper bound remains
bounded for a fixed window-initial error.
When $\kappa=1$, this reduces to
\begin{equation}
d_{m,t}
\leq
d_{m,t_s}+(t-t_s)(\xi+\Delta),
\label{eq:supp_linear_error_accumulation}
\end{equation}
showing that the tracking-error upper bound grows at most
linearly with the length of continuous fast-parameter transfer.
When $\kappa>1$, Eq.~\eqref{eq:supp_window_error_unrolling}
permits geometric growth of the worst-case bound.
These upper bounds do not imply that the actual tracking
error necessarily increases. Nevertheless, without periodic
policy anchoring, continuous fast-parameter transfer may
allow historical errors to accumulate or amplify, increasing
the risk of error cascades, particularly when $\kappa\geq1$.

\noindent\textbf{Illustrative Example of Unbounded Error Accumulation.}
To illustrate that unbounded tracking-error accumulation
is possible, consider a one-dimensional adaptation process
over an arbitrarily long sequence without periodic anchoring.
Let $\mathcal{U}_t(x)=x$, the ideal parameter be
$x_t^{\star}=0$, and the adaptation perturbation be a
constant $\epsilon>0$. Starting from $x_0=0$, the
perturbed adaptation satisfies

\begin{equation*}
x_{t+1}=x_t+\epsilon,
\qquad x_t=t\epsilon.
\end{equation*}
Consequently, the tracking error is
$d_t=|x_t-x_t^{\star}|=t\epsilon$, which grows
without bound as $t\to\infty$.
This construction satisfies Assumptions~1--3 with
$\xi=\epsilon$, $\Delta=0$, and $\kappa=1$.
It demonstrates that unbounded error accumulation
is possible under the stated assumptions.

\noindent\textbf{Corollary 1 (Window-confined Error Propagation).}
Let $w=|\mathcal{W}_m|=t_e-t_s+1$ denote the size of temporal window
$\mathcal{W}_m$. At the beginning of each window, DISRL initializes the
snapshot-level fast parameter from the window-level master parameter, i.e.,
$\Phi_{m,t_s}\leftarrow\Theta_m$. Since the fast parameter is directly
transferred only within the current window, the recurrence in
Eq.~\eqref{eq:supp_window_error_unrolling} can be expanded for at most
$w-1$ consecutive transition steps.

When $\kappa=1$, the window-end error is therefore bounded by
\begin{equation}
d_{m,t_e}
\leq
d_{m,t_s}+(w-1)(\xi+\Delta).
\label{eq:supp_linear_window_bounded_error}
\end{equation}

This bound highlights the role of periodic
master-policy anchoring in limiting error
propagation within each temporal window.
Specifically, the additional error contribution
from continuous fast-parameter transfer is bounded
by $(w-1)(\xi+\Delta)$ when $\kappa=1$, while
the window-initial error
$d_{m,t_s}=\|\Theta_m-\Phi_{m,t_s}^{\star}\|_2$
is determined by the evolving master parameter. This window-level bound does not imply a uniform
error bound across windows, as the master-parameter
initialization error remains unbounded by the analysis.

In contrast, without window-level anchoring, continuous fast-parameter
transfer over the entire graph sequence would make the direct accumulation
length grow with the total number of snapshots. Since the window size $w$
is fixed and smaller than the full sequence length $T$, the dual-scale
design restricts direct error propagation to each
temporal window through periodic master-policy anchoring,
thereby mitigating the risk of unbounded long-horizon drift.

\section{Evaluation Settings and Hyper-parameters}
\label{sec:supp_evaluation_settings}

\subsection{Dynamic Generation Procedures}

We implement four core topological operators alongside continuous background
noise (random node/edge insertions and deletions) to evolve graphs into
temporal snapshot sequences:
\begin{itemize}
    \item \textbf{Expansion \& Shrinkage:} Expansion injects boundary nodes
    mimicking local homophily connectivity patterns, whereas shrinkage prunes
    peripheral members while preserving the structural core.
    \item \textbf{Merging \& Splitting:} Merging introduces bridge edges
    between structurally related community pairs to unify their labels,
    whereas splitting fractures a cohesive group into disjoint, internally
    coherent sub-components.
\end{itemize}

\subsection{Evaluation Protocols}

We evaluate simulated evolution under two scenarios and additionally use a real-world dynamic dataset:
\begin{itemize}
    \item \textbf{Expansion \& Shrinkage Dynamic Setting:} This setting is
    evaluated on the four SNAP datasets. Starting from $t=1$, Amazon
    and DBLP undergo six consecutive expansion steps followed by six
    shrinkage steps, and then twelve additional steps in which expansion
    or shrinkage is randomly selected at each step, producing $T=25$
    snapshots. YouTube and Twitter undergo six consecutive expansion
    steps followed by six shrinkage steps, producing $T=13$ snapshots.
    We randomly sample 100 target seed nodes from each of the Amazon,
    DBLP, YouTube, and Twitter datasets.

    \item \textbf{Merge \& Split Dynamic Setting:} This setting is evaluated
    on Amazon and DBLP due to their sharp modularity boundaries. Each
    sequence undergoes four merge operations followed by four split
    operations, producing $T=9$ snapshots over 100 sampled target seed nodes
    per dataset.

    \item \textbf{Real-world Dynamic Setting:} Football is derived from the
    Century of College Football dataset~\cite{century-college-football-dataset}.
    We aggregate games from 2000--2024 into 24 yearly snapshots, excluding
    2020 due to the COVID-19 pandemic, and use each team's annual conference
    affiliation as its community label. We track 10 fixed target teams
    present in all snapshots, without introducing any artificial
    community evolution events.

\end{itemize}

\begingroup
\noindent
The query-specific supervision exclusion is performed
independently at each snapshot by the evaluation
procedure. The target community's ground-truth annotation
is used to identify and remove it from the candidate
supervision pool and to compute evaluation metrics,
but is not supplied as training supervision for that
query. For each independently tracked seed, the fast policy
is adapted using the available annotations before
target prediction. Upon the arrival of the next
snapshot, its permitted supervision is used to compute
transition meta-gradients, which are buffered rather
than immediately applied. These gradients are aggregated
to update the master policy only at window boundaries,
without accessing unobserved future snapshots.
\par
\endgroup

\subsection{Hyper-parameter Configurations}

Table~\ref{tab:disrl_hyperparameters} summarizes the key hyper-parameter
configurations used in the experiments. 

\begin{table}[!htbp]
\centering
\caption{Hyper-parameter configurations of DISRL.}
\label{tab:disrl_hyperparameters}
\begingroup
\scriptsize
\setlength{\tabcolsep}{2.7pt}
\renewcommand{\arraystretch}{0.98}
\begin{tabular*}{\columnwidth}{@{\extracolsep{\fill}}L{0.27\columnwidth}L{0.43\columnwidth}L{0.20\columnwidth}@{}}
\toprule
\textbf{Component} & \textbf{Configuration} & \textbf{Value} \\
\midrule
\multirow{2}{*}{Supervision Retrieval}
& Known communities & 100,all except target  \\
& Top-$K$ & 50,5 \\
\midrule
\multirow{7}{*}{Community Explorer}
& Batch size & 16 \\
& Epochs per snapshot & 50 \\
& Hidden dimension & 64 \\
& iGPN layers & 3,2 \\
& MLP layers & 2 \\
& Learning rate & $10^{-2}$ \\
& Discount factor $\gamma$ & 0.99 \\
\midrule
\multirow{3}{*}{Dual-scale Update}
& Window size & 4 \\
& Decay factor $\rho$ & 0.8 \\
& Learning rate $\eta$ & 0.01 \\
\bottomrule
\end{tabular*}
\endgroup
\end{table}

\FloatBarrier

\section{Case Study}
\label{sec:supp_case_study}

\begin{figure}[!htbp]
\centering
\includegraphics[width=0.90\columnwidth]{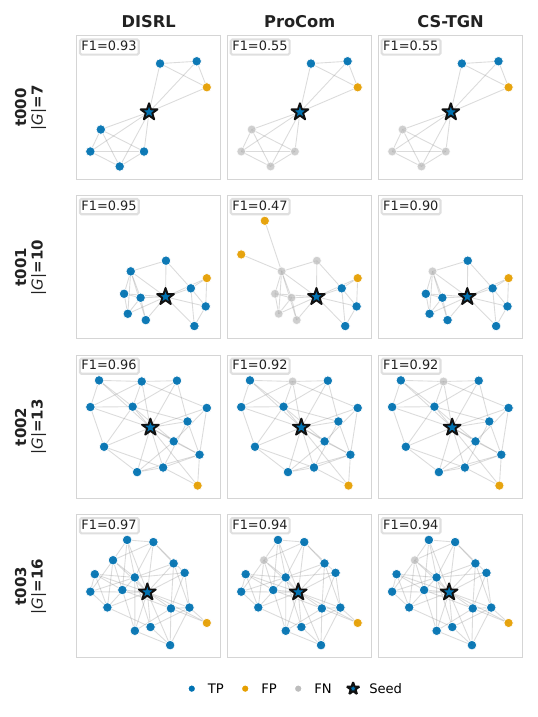}
\caption{DBLP case study over four snapshots. Rows show snapshots and columns show methods; each panel reports F1. The target community grows from 7 to 16 nodes.}
\Description{DISRL, ProCom, and CS-TGN predictions over four DBLP snapshots, with true positives, false positives, false negatives, and the target seed identified.}
\label{fig:case_study}
\end{figure}

To visually dissect the boundary tracking behavior of DISRL under
fine-grained topological mutations, Fig.~\ref{fig:case_study} qualitatively
orchestrates a temporal case study on DBLP under expansion\&shrinkage
dynamics. The tracking sequence tracks a specific target seed node $s$
across four consecutive snapshots ($t_0$ to $t_3$), during which the local
ground-truth community size dynamically swells from $7$ to $16$ nodes.

Empirically, DISRL demonstrates remarkable adaptive stabilization throughout
this structural expansion, registering high F1-scores of $0.93$,
$0.95$, $0.96$, and $0.97$ chronologically. In sharp contrast, both
\textit{ProCom}~\cite{wu-2024-procom} and
\textit{CS-TGN}~\cite{hashemi-2023-cstgn} suffer from severe synchronization
delay at the initial onset ($t_0$), stagnating at an F1-score of
$0.55$. As the stream advances to $t_1$, \textit{ProCom} fails to capture
the newly incorporated members and its tracking accuracy drops to $0.47$.
Concurrently, while the dynamic GNN baseline \textit{CS-TGN} slowly
recaptures the community boundary through continuous representation updates,
its accuracy consistently trails DISRL across the entire temporal window.

This qualitative divergence provides a clear microcosm of the underlying
algorithmic mechanisms. The early-stage performance collapse of
\textit{ProCom} and \textit{CS-TGN} uncovers their vulnerability to
cold-start topological changes; without specialized localized historical
anchors, they easily over-expand into adjacent dense graph fragments.
Conversely, DISRL circumvents this tracking lag via its dual-stage
coordination. At the initial frame $t_0$, the cold-start bootstrapping policy
isolates the immediate vicinity of seed $s$. In subsequent snapshots
($t_1 \rightarrow t_3$), the reference community synthesis operator
explicitly applies topological filtering through the reference-synthesis
rule defined in the main paper to prune disconnected structural noise. This
mechanism allows the explorer agent to seamlessly inherit previous
predictions as historical context, accurately identifying boundary-expanding
members while preserving temporal tracking continuity.

\section{Efficiency Study}
\label{sec:supp_efficiency}

\begin{figure}[!htbp]
\centering
\includegraphics[width=\columnwidth]{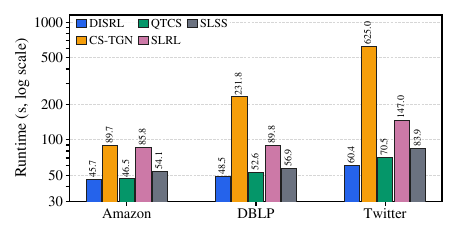}
\caption{Efficiency comparison under the expansion\&shrinkage setting. The
reported runtime is normalized per target seed per snapshot over 50
independent queries.}
\Description{Running times for DISRL, CS-TGN, QTCS, SLRL, and SLSS on Amazon, DBLP, and Twitter, plotted on a logarithmic scale.}
\label{fig:efficiency_study}
\end{figure}

To evaluate the computational efficiency of DISRL, we benchmark its
wall-clock execution time against competitive baselines under the
expansion\&shrinkage setting. As illustrated in
Fig.~\ref{fig:efficiency_study}, DISRL consistently outperforms all
baselines, achieving the lowest execution latency across all three datasets.
Compared with CS-TGN, DISRL obtains $1.96\times$, $4.78\times$, and
$10.35\times$ speedups, respectively. The advantage is especially clear on
Twitter, where the larger and denser topology increases the cost of
repeatedly processing temporal graph structures. The efficiency difference
comes from the execution patterns of the compared methods. CS-TGN learns
temporal GNN representations over all snapshot graphs and repeatedly scans
training queries across epochs. SLRL retrains a local detector for each
seed-snapshot pair, where each detector performs two optimization rounds and
repeatedly updates the expander across multiple epochs. SLSS incurs costly
similar-community matching during community expansion, while QTCS is slowed
by seed-wise TPPR-guided ALS search over temporal transition states. In
contrast, DISRL performs policy adaptation on a compact Top-$K$ seed-specific
relevant supervision set and localized subgraphs, avoiding full-graph
temporal representation learning.

\FloatBarrier

\end{document}